\documentclass[letterpaper, 10 pt, conference]{ieeeconf}  
\IEEEoverridecommandlockouts                              

\usepackage{placeins}
\usepackage{amsmath,amssymb,amsfonts}
\usepackage{algorithmic}
\usepackage{graphicx}
\usepackage{textcomp}
\usepackage{siunitx}
\usepackage{hyperref}
\usepackage{mathtools}
\usepackage{nicefrac}
\usepackage{booktabs}

\usepackage{xcolor}
\usepackage{lipsum}

\DeclareMathOperator{\atantwo}{atan2}
\DeclareMathOperator{\lift}{lift}

\title{\LARGE \bf Globally Optimal Inverse Kinematics as a Non-Convex Quadratically Constrained Quadratic Program*}

\author{Tomáš Votroubek$^{1}$ and Tomáš Kroupa$^{2}$
\thanks{*The authors acknowledge the support of the OP VVV funded project CZ.02.1.01/0.0/0.0/16\_019/0000765 ``Research Center for Informatics''. This research was partly funded by AFOSR award FA9550-18-1-0097.}
\thanks{$^{1,2}$Kroupa and Votroubek are with Faculty of Electrical Engineering, Czech Technical University in Prague, Charles Square 13, Prague 2, Czech Republic {\tt\small tomas.votroubek@fel.cvut.cz} and {\tt\small tomas.kroupa@fel.cvut.cz}}%
}

\begin{document}

\maketitle
\thispagestyle{empty}
\pagestyle{empty}

\begin{abstract}
We show how to compute globally optimal solutions to inverse kinematics by formulating the problem as a non-convex quadratically constrained quadratic program. Our approach makes solving inverse kinematics instances of generic redundant manipulators feasible. We demonstrate the performance on randomly generated designs and real-world robots with up to ten revolute joints. The same technique can be used for manipulator design by introducing kinematic parameters as variables. 
\end{abstract}

\section{INTRODUCTION}
\emph{Inverse kinematics} (IK) is the problem of finding control parameters that bring a robot into a desired pose within its kinematic constraints. For manipulators of six or fewer joints, this problem is practically solvable and has at most 16 different solutions \cite{Manocha}. Adding additional joints results in an under-constrained problem with infinitely many feasible solutions satisfying the pose constraints. Therefore, it is interesting not simply to look for feasible solutions but optimal ones. In this paper, we focus on minimizing the total movement of joints. Specifically, the task is as follows: Given the Denavit-Hartenberg parameters \cite{Spong2020-mt} of a kinematic chain, input joint angles, and a desired pose for its end effector, find joint angles that achieve the desired pose while minimizing the weighted sum of differences from the input (preferred) angles.

The presented problem is very hard in general. The current state-of-the-art technique is based on \emph{sum of squares} (SOS) optimization and Gröbner bases, which are both NP-hard in the context of multivariate polynomial equations. Trutman \textit{et al.} demonstrate \cite{trutman2022globally} that for manipulators with at most seven joints, the optimal solutions admit a quadratic SOS representation, thereby being practically solvable. Such results do not extend to manipulators with more than seven joints.

A common way to avoid the complexity of IK is either to find an approximate solution or to design the robotic arm to be simple to control. 
Approaches from the first category include an approximation
based on the Jacobian by Buss \cite{Buss_undated-fo}, which uses numerical methods to find local optima. Similarly, Dai \textit{et al.} \cite{Dai2019-uf} designed a global solution method for the convex approximation of the original IK. On the other hand, the design of the \href{https://www.kuka.com/en-us/products/robotics-systems/industrial-robots/lbr-iiwa}{KUKA LBR iiwa} manipulator falls into the second category. However, having to design around computational constraints has the downside of limiting the achievable performance of a manipulator.

The problem may be further relaxed for real-time control by dropping the guarantees on optimality or convergence to the desired pose. For example, \href{https://robotology.github.io/robotology-documentation/doc/html/group__iKinSlv.html}{iKinSlv} finds a locally optimal solution which minimizes the distance from the desired pose using interior point optimization; the closed-loop inverse kinematics algorithm suggested in \href{https://github.com/stack-of-tasks/pinocchio/blob/97a00a983e66fc0a58667f9671e2d806ba9b730b/examples/inverse-kinematics.cpp}{the Pinocchio library} attempts to iteratively reduce the pose error, while modern approaches \cite{ExampleQP} based on convex \emph{quadratic programming} (QP) rely on asymptotic convergence.

Global solvers recover a globally optimal solution whenever it exists or guarantee that none exists. In the context of mechanical design, it is possible to optimize a design by computing and comparing the actual limits of performance in some desired metric specified by the user. Using the optimality guarantees to test existing solvers' properties is also possible. If a solver fails to provide a solution, it is possible to answer whether the instance was solvable.

The contribution of this paper is based on the premise that while SOS optimization is, in some sense, the correct answer to the \emph{polynomial optimization problem} (POP), incumbent implementations fall significantly short of their promised potential. We will show that highly optimized off-the-shelf global optimizers can converge faster for problems such as IK despite having much higher computational complexity. Instead of reformulating IK into a theoretically easier problem, we reformulate it into a problem for which efficient solvers exist.

Our approach finds globally optimal solutions of generic seven \emph{degree-of-freedom} (DOF) IK instances faster than the state-of-the-art. Our technique is unique in finding globally optimal IK solutions for generic manipulators with eight or more degrees of freedom.

\section{QUADRATIC PROGRAM FORMULATION}
Starting with the POP for globally solving IK recently proposed by Trutman \textit{et al.} \cite{trutman2022globally}, we lift it into a \emph{quadratically constrained quadratic program} (QCQP). Specifically, we find a new basis in lifting variables of the feasible set of the initial POP, such that the new higher-dimensional representation is at most quadratically constrained. For example, for a polynomial constraint $x_1x_2x_3=0$ we introduce a variable $y$ and rewrite the constraint as $y=x_1x_2$ and $yx_3=0$. Solvers for non-convex QCQPs, such as \href{www.scipopt.org/}{SCIP} \cite{scip} or \href{www.gurobi.com/}{Gurobi} \cite{gurobi}, can find the global solution to this new problem by the spatial branch-and-bound method.

\subsection{Denavit–Hartenberg parameters}
\label{DH}
It is conventional to describe $n$-joint manipulators in terms of their \emph{Denavit–Hartenberg} (D-H) parameters for each link $i=1,\dots,n$: the offset $d_i$, the link length $r_i$, the link twist $\alpha_i$, and the joint angle variable~$\theta_i$.

\begin{figure}[htpb]
    \center
    \includegraphics[width=0.6\linewidth]{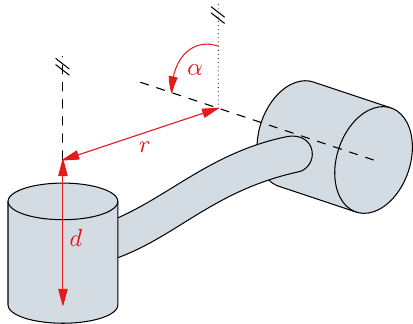}
    \caption{Diagram of Denavit–Hartenberg parameters}
\end{figure}
\noindent
The transformation between successive coordinate frames from link $i$ to $i-1$ is defined by a D-H matrix $T_i(\theta_i) =$
\[
\left[ {\begin{array}{rrrr}
 \cos\theta_i & -\cos\alpha_i\sin\theta_i &  \sin\alpha_i\sin\theta_i & r_i\cos\theta_i  \\
 \sin\theta_i &  \cos\alpha_i\cos\theta_i & -\sin\alpha_i\cos\theta_i & r_i\sin\theta_i  \\
              0 &                \sin\alpha_i &                \cos\alpha_i &               d_i  \\
              0 &                             0 &                             0 &                 1  \\
\end{array} } \right].
\]
The end effector pose w.r.t. the base coordinate system is represented by the matrix $P$, which induces the \emph{pose constraint} $T_1(\theta_1)\dotsb T_n(\theta_n)= P$.

\subsection{Polynomial Formulation}
In this section we describe the POP for IK initially formulated by Trutman \textit{et al.} \cite{trutman2022globally}. This formulation avoids trigonometric functions in the D-H matrices. For each link $i=1,\dots,n$ introduce variables $c_i \coloneqq \cos \theta_i $ and $s_i\coloneqq \sin \theta_i$, along with the identity $c_i^2 + s_i^2 = 1$. After the change of variables, every D-H matrix $T_i(\theta_i)$ depends only on $c_i$ and~$s_i$, and we denote is simply by $T_i$.  We assume that each joint angle $\theta_i$ is constrained by the maximal value $\theta^{\mathit{max}}_i\in [0,\pi]$ and the minimal value $\theta^{\mathit{min}}_i\in [-\pi,0]$. The vector of preferred joint angles is $(\hat{\theta}_1,\dots,\hat{\theta}_n)$ and we consider some weights $w_1,\dots,w_n\ge 0$ satisfying $w_1+\dotsb + w_n=1$. The~goal is to minimize the weighted distance 
\begin{equation}\label{criterion}
\sum_{i=1}^n w_i\bigl((c_i-\cos \hat{\theta}_i)^2+(s_i-\sin \hat{\theta}_i)^2\bigr).
\end{equation}
The choice of preferred joint angles $(\hat{\theta}_1,\dots,\hat{\theta}_n)$ makes it possible to minimize the total movement of joints, or to find configurations where some links achieve desirable orientations (e.g. an upright posture in humanoid robots \cite{minimumjerk}).

Further, we split the kinematic chain with $n$ joints into two subchains that meet in the middle\footnote{The placement of the split does not significantly affect the runtime as long as each subchain involves at least two joints.}: the first anchored at the origin $1$ and ending at $\nu$ and the second starting at $\nu+1$ and ending at $n$. The pose constraint $T_1(\theta_1)\dotsb T_n(\theta_n)= P$ then becomes $T_1\dotsb T_\nu = P T_n^{-1}\dotsb T_{\nu+1}^{-1}$. 
This formulation enables us to lower the degree of polynomials appearing in the above constraint and, consequently, to reduce the number of lifting variables (see Section \ref{sec:lift}). 

The resulting POP (see \cite[(9)]{trutman2022globally}) with variables $\mathbf{c}=(c_1,\dots,c_n)$, $\mathbf{s}=(s_1,\dots,s_n)$, the objective function \eqref{criterion}, and the constraints formulated above is:
\begin{alignat}{3}
\min_{\mathbf{c}, \mathbf{s}} &\quad& \sum_{i=1}^n 2w_i(1-c_i\cos&\,\hat{\theta}_i-s_i\sin\hat{\theta}_i) \label{linobj}    \\
\text{s.t.}   &\quad&     P T_n^{-1}\dotsb T_{\nu+1}^{-1} &= T_1\dotsb T_\nu \label{polycons}\\
              &\quad& (c_i + 1) \tan \frac{\theta^{\mathit{min}}_i}{2} & \le s_i &i=1,\dots,n \label{min}\\
              &\quad& (c_i + 1) \tan \frac{\theta^{\mathit{max}}_i}{2} & \ge s_i &i=1,\dots,n \label{max}\\
              &\quad&                                    c_i^2 + s_i^2 &  =  1   &i=1,\dots,n \label{popfinal}
\end{alignat}
Since \eqref{criterion} is equal to the linear function in  \eqref{linobj}, this POP is in fact a minimization problem with a linear objective and non-convex polynomial constraints. Linear inequalities \eqref{min}-\eqref{max} are equivalent to design constraints $\theta^{\mathit{min}}_i\le \theta_i \le \theta^{\mathit{max}}_i$ by basic trigonometric identities and the convention $0\le\theta^{\mathit{max}}_i\le \pi$,  $-\pi\le\theta^{\mathit{min}}_i\le 0$. We note that the last set of constraints from the original POP \cite[(9)]{trutman2022globally} is redundant.

We use $\theta_i = \atantwo(s_i, c_i)$ to recover the optimal joint angles $\theta_i$. In the next section, we show how to make the POP above into an equivalent QCQP, that is, how to reduce the degree of polynomials in the constraints \eqref{polycons}.

\subsection{Lifting Algorithms}\label{sec:lift}
The SOS-based method \cite{trutman2022globally} to solve the non-convex POP \eqref{linobj}--\eqref{popfinal} uses the hierarchy of moment relaxations \cite{lashierarchy}, which are convex optimization problems and whose values converge to the optimal value of the initial POP. Thus, instead of optimizing a POP directly, the relaxations optimize its linearization using semidefinite programming. This is based on the following convexification idea. For a function $p$ over~$X$, the minimizer of the expected value $\int_X p(x)d\mu(x)$ is a probability measure $\mu$ supported by the minimizers of function $p$. In practice, this approach iteratively attempts to represent probability measures using truncated pseudo-moment sequences. However, to obtain a global solution rather than a relaxation, it is sufficient to constrain the first moment matrix of $\mu$ to be rank one \cite{Laurent2009}. This constraint can be encoded in a  QCQP and ensures that $\mu$ is a Dirac measure, which guarantees the solution's global optimality. On the one hand, this approach is computationally expensive as it introduces an excessive number of variables and many redundant constraints even after picking an appropriate basis. On the other hand, the computational methods for POPs based on hierarchies of convex relaxations are certifiably globally optimal. 

As an alternative, we approach the problem from the opposite direction, in a way that exploits the problem's sparsity and is also easier for a presolver to optimize. We rewrite each polynomial constraint of the POP into a set of equivalent quadratic constraints in a higher dimensional space. That is, we rewrite the monomials of each polynomial constraint by repeatedly substituting the multiplication of two optimization variables with a new lifting variable equal to their product. For example, the monomial $x_1x_2x_3\cdots x_n$ could be lifted as $y_{12}x_3\cdots x_n$, where $y_{12}=x_1x_2$. Each application of such a rewriting reduces the degree of the polynomial and can be applied repeatedly to each monomial of the original expression until only single variables remain. Since the lifting variables are quadratically constrained, we effectively transform the non-convex POP \eqref{linobj}--\eqref{popfinal} into a non-convex QCQP. Such a straight-forward lifting technique was already discussed by Shor \cite{Shor1987}.

Due to the structure of the pose constraint, we can apply the same rewriting to the expressions involving D-H matrices directly. Since the result of $T_iT_j$ will contain at most quadratic terms, we can introduce a matrix of lifting variables for every matrix product and proceed as described above.

The number of variables and constraints in the resulting program depends on the order of application of such a rewriting rule. For example, the constraint $x_1x_2x_3x_4 + x_2x_3x_5 = 0$ can be reformulated in at least two different ways:

\vspace{0.5em}
\noindent\begin{minipage}{.5\linewidth}
\[
\label{eq:liftA}
\tag{A}
\begin{aligned}
    0 &= y_{12}y_{34} + y_{23}x_5 \\
    y_{12} &= x_1x_2 \\
    y_{34} &= x_3x_4 \\
    y_{23} &= x_2x_3 \\
\end{aligned}
\]
\end{minipage}%
\noindent\begin{minipage}{.5\linewidth}
\[
\tag{B}
\begin{aligned}
    0 &= y_{123}x_4 + y_{23}x_5 \\
    y_{23} &= x_2x_3 \\
    y_{123} &= x_1y_{23} \\
\end{aligned}
\]
\end{minipage}
\vspace{0.5em}

\noindent Taking advantage of repeated terms, and even the order in which products are lifted, has an observable impact on the speed and behavior of solvers and on the cuts that they apply. 

In the next section, we will focus on the performance of the method corresponding to System~\ref{eq:liftA} and the matrix based method. We will refer to these methods as $\lift_\text{A}$ and $\lift_\text{M}$, respectively. While $\lift_\text{M}$ often results in the fewest variables after presolve, it is not always the fastest. We found the behavior of $\lift_\text{A}$ more predictable. We refer the interested reader to our \href{https://github.com/votroto/IK.jl}{code repository} for the implementation details.

\subsection{Global Optimality Certificates}
The global optimality certificate for the solution provided by the SOS-based method \cite{trutman2022globally} is based on evaluating the rank of the matrix of floats returned by the solver.

\emph{Branch and bound} (BB) algorithms are global optimization methods. They maintain a provable upper and lower bound on the (globally) optimal objective value, and they return a certificate proving that the solution found is globally optimal within a predefined margin of error. Our implementation involves the solver Gurobi, which uses a variant of the spatial BB method. Gurobi does not return a certificate of global optimality. The found optimal solution can be used to confirm Gurobi's final primal bound. The log output contains the BB tree information, but it seems impossible to completely reconstruct Gurobi's BB tree or view the specific added cuts. 

Admittedly, using Gurobi means a certain trade-off between certifying the global optimality and efficiency of the computations. In our experiments, we recovered the same global solutions as in \cite{trutman2022globally} and scaled our algorithm to manipulators with more DOFs than other research papers. Recent results \cite{dey2021branch} indicate that considering random instances of some problems results in polynomial time complexity for BB methods.

\subsection{Applicability of QCQP Lifting}
The lifting algorithm presented in Section \ref{sec:lift} applies to any IK problem involving polynomials since it needs no additional assumptions about the degree of polynomials or their properties. This enlarges the scope of applicability of our model significantly. For example, instead of the distance from preferred angles \eqref{criterion}, we can optimize \emph{manipulability}. We can also use the QCQP lifting for the \emph{design of manipulators} by introducing new variables in place of the parameters of the manipulator. This is based on the observation that even if all the parameters in the D-H matrices are treated as optimization variables, the resulting IK problem would still be a POP. Since polynomials are closed under addition and multiplication, the resulting constraints could be lifted as described in Section \ref{sec:lift}.

\section{RESULTS}
We compare our QCQP-based procedure to the prior SOS-based method \cite{trutman2022globally}. Our numerical experiments use an \emph{Intel Xeon Gold 6146} processor limited to 4 threads. Implementing the SOS-based method involves Maple 2022 and Mosek~10, while our method uses Gurobi 10. These solvers are the latest and fastest for each problem class available to us at the time of writing. Results for the SOS-based method are based on the original implementation and use all the techniques described in \cite{trutman2022globally} to improve numerical stability. We show the statistics of solve times estimated from randomly generated feasible and infeasible poses of the \href{https://www.kuka.com/en-us/products/robotics-systems/industrial-robots/lbr-iiwa}{KUKA LBR iiwa} 7-DOF manipulator, the \emph{iCub} \cite{icub} humanoid robot's right arm and torso (7--10 DOF), and on randomly generated designs with seven joints. Unless stated otherwise, the  results in following sections pertain to the lifting approach $\lift_\text{A}$ (see Section \ref{sec:lift}).

\subsection{Factors Influencing the Solution Time}
As we solve IK via a non-convex QCQP, it is difficult to predict the solution time of a given instance; however, in our experience, the time increases with the number of non-zeros in the constraint matrix, the number of variables, and the volume of the space that needs to be searched. In practical terms, the runtime depends on the manipulator's range of motion and the link twists. Specifically, designs with axes offset by $90^\circ$, as is the case for the \emph{iCub} and the \emph{KUKA~LBR~iiwa}, are easier to solve than ones with joints at arbitrary angles. 

To demonstrate how the time required to solve IK depends on the aspects of mechanical design, we will consider three sets of randomly generated robotic designs with seven joints:
\begin{description}
    \item[Orth.] (Orthogonal) designs with link twists sampled uniformly from $\{\nicefrac{-\pi}{2}, \nicefrac{\pi}{2}\}$, resulting in orthogonal axes and a 6-radian range of motion $-3 \leq \theta_i \leq 3$, 
    \item[6-rad] designs with arbitrary link twists $\alpha_i$ uniformly sampled from $[-3,3]$ and  6-radian range of motion,
    \item[4-rad] designs with arbitrary twists $\alpha_i$ uniformly sampled from $[-3,3]$, and  4-radian range $-2 \leq \theta_i \leq 2$.
\end{description}
The link offsets and lengths are generated randomly according to a uniform distribution between $10$ and $100$ centimeters.

Finally, warm-starting can speed up the global solution search even with low-precision local solutions. We used \emph{IPopt} \cite{ipopt} and the straightforward non-linear program \ref{localprog} to find an initial guess:
\begin{equation}
\label{localprog}
\begin{aligned}
\min_{\boldsymbol{\theta}} &\quad&& \sum_{i=1}^n 2w_i(1-\cos\theta_i\cos\,\hat{\theta}_i-\sin\theta_i\sin\hat{\theta}_i)     \\
\text{s.t.}   &\quad&&    T_1\dotsb T_\nu = P T_n^{-1}\dotsb T_{\nu+1}^{-1}   \\
              &\quad&&   \theta^{min}_i \leq \theta_i \leq \theta^{max}_i   \hspace{4em} i=1,\dots,n
\end{aligned}
\end{equation}
This warm-starting may fail to provide a solution, as local solvers may get stuck at points of local infeasibility. We limited the local-solution search to 200 iterations (approximately $\qty{200}{\milli\second}$). Unless indicated otherwise, all of the following benchmarks used warm-starting. 

Switching to modern QP or iterative real-time techniques with higher precision and lower failure rates will likely improve solution times. Additionally, such techniques could be used as runtime branch-and-bound heuristics. 

\begin{table}
\renewcommand{\arraystretch}{1.3}
\caption{QCQP runtime, the positional and rotational error of the final solution. ``{Cold}'' is a cold-start without a local solution. The $\lift_\text{M}$ formulation solved the orthogonal instances in $\qty{1.23}{\second}$, \emph{4-rad} in $\qty{1.93}{\second}$, but \emph{6-rad} in $\qty{8.96}{\second}$  on average.}
\label{table:randomresults}
\centering
\begin{tabular}{r||rrr}
\hline
\bfseries{Set} & \bfseries{Time ($\si{\second}$)} & \bfseries{Err. ($\si{\micro\meter}$)} & \bfseries{Err. ($\si{\micro\radian}$)} \\
\hline
{Orth.}        & 1.31 & 2.29 & 1.28 \\
{(cold) Orth.} & 2.12 & 2.71 & 1.28\\
{4-$\si{\radian}$}        & 2.36 & 2.20 & 1.09\\
{(cold) 4-$\si{\radian}$} & 3.90 & 2.39 & 1.26\\
{6-$\si{\radian}$}        & 4.91 & 2.88 & 1.70\\
{(cold) 6-$\si{\radian}$} &12.20 & 2.51 & 1.15\\
\hline
\end{tabular}
\end{table}

\begin{figure}[tpb]
  \includegraphics[width=\linewidth]{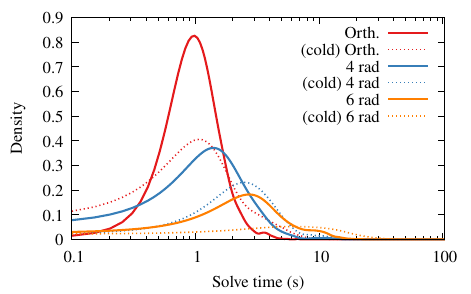}
    \caption{A semi-log plot of solve-time kernel-density-estimates from 1000 samples of randomly generated designs for each parameter set.}
\end{figure}

Note that the variable and constraint count alone are not a good predictor of runtime. Despite not allowing for cancellation $\lift_\text{M}$ results in the fewest variables after presolve; however, while the method is faster on some designs on average, its behavior is less predictable. In extreme cases, this resulted in runtimes almost 100 times higher than the average. 

\begin{table}
\renewcommand{\arraystretch}{1.3}
\caption{{DOF} and the resulting problem size after presolve for each lifting method}
\label{tab:progsize}
\centering
\begin{tabular}{rr||rr|rr}
    \hline
    \multicolumn{2}{c||}{} & \multicolumn{2}{c|}{$\mathbf{\lift_\text{M}}$} & \multicolumn{2}{c}{$\lift_\text{A}$}\\
    \hline
    \bfseries{Set}  & \bfseries{\textsc{dof}} & \bfseries{vars.} & \bfseries{constrs.}  & \bfseries{vars.} & \bfseries{constrs.} \\
    \hline
    {KUKA}  &   7 &   71 &  57 &  76 &  62 \\
    {Orth.} &   7 &   78 &  64 &  80 &  66 \\
    {6-$\si{\radian}$} &   7 &  118 & 104 & 119 & 105 \\
    {iCub}  &   7 &   78 &  62 &  78 &  64 \\
    {iCub}  &   8 &   95 &  79 & 102 &  86 \\
    {iCub}  &   9 &  136 & 118 & 146 & 128 \\
    {iCub}  &  10 &  179 & 159 & 190 & 170 \\
    \hline
\end{tabular}
\end{table}

\subsection{Comparison Across Solvers and Methods (7 DOF)}
\begin{figure}[tpb]
  \includegraphics[width=\linewidth]{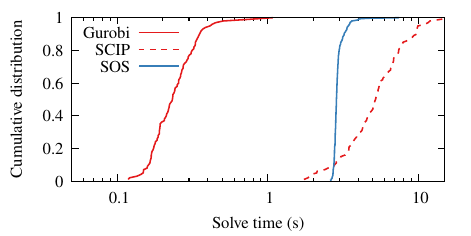}
  \caption{A semi-log plot comparing the normalized cumulative distributions from 100 solve times for the \emph{KUKA LBR iiwa} 7-DOF robot using our QCQP approach (Gurobi, SCIP) and the SOS-based technique.}
  \label{fig:compare.kuka}
\end{figure}

Our approach outperforms the SOS-based method \cite{trutman2022globally} when tested on feasible IK instances generated by the \emph{KUKA LBR iiwa} 7-DOF manipulator. The state-of-the-art method was explicitly optimized to take advantage of the manipulator's structure and eliminate one of the joints. Due to Maple's license availability, the following comparison was performed on four threads of \emph{AMD Ryzen 5 4600G}. We did not include the modelling and Matlab overhead in the reported runtimes. The average solve-time (Figure \ref{fig:compare.kuka}) of our technique was $\qty{0.26}{\second}$, compared to $\qty{2.9}{\second}$ for the optimized SOS approach. We have also tried the free and open-source solver \emph{SCIP} instead of \emph{Gurobi}. However, the average solve time for \emph{SCIP} rises to $\qty{5.7}{\second}$. While a performance decrease is expected when comparing commercial solvers to free and open-source ones, in this case \emph{SCIP} could also not effectively utilize more than one thread and did not benefit from the local solution provided by warm-starting. The prior method can take advantage of the structure of the manipulator, which significantly improves the runtime compared to generic manipulators. 

\begin{figure}[tpb]
	\includegraphics[width=\linewidth]{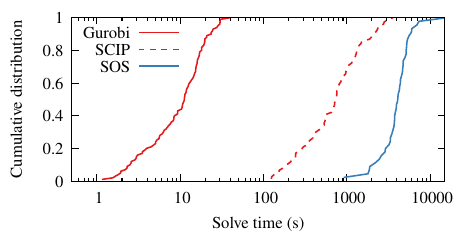}
	\caption{A semi-log plot comparing the normalized cumulative distributions of solve times obtained from 100 samples of randomly generated designs with 6-radian range of motion using our approach and the SOS-based technique (rounding to two significant digits).}
	\label{fig:compare.rand}
\end{figure}

To compare the methods on randomly generated designs, we rounded the lengths and offsets of the \emph{6-rad} test-set to two significant digits, as the performance of the SOS-based method depends on the number of digits in the problem. The performance of our method is not significantly affected by the rounding. The average solve-time (Fig.~\ref{fig:compare.rand}) of our technique was $\qty{12.0}{\second}$ with \emph{Gurobi} and $\qty{931}{\second}$ with \emph{SCIP}. The modified SOS approach took $\qty{4328}{\second}$ to solve an instance on average. Finally, we note that the prior SOS-based method has roughly a $1\%$ failure rate, a problem we did not encounter with our method.

\begin{figure}[tpb]
	\includegraphics[width=\linewidth]{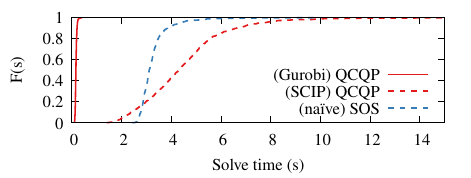}
	\caption{Normalized cumulative distributions $F(s)$ of solve times from 1000 samples of the \emph{iCub's} 7-DOF right arm. The full SOS-based method with symbolic reduction did not finish in time.}
	\label{fig:compare.icub}
\end{figure}

The symbolic reduction step of the SOS-based method can be a bottleneck for some robots. For example, the 7-DOF IK instances of \emph{iCub}'s right arm (Fig.~\ref{fig:compare.icub}) did not consistently finish within 5 hours of runtime even despite rounding to the nearest millimeter (3 significant figures). In this case however, we found that the reduction is not necessary, and \emph{Mosek} could solve the SOS programs directly with only a $1\,\%$ failure rate (this was described as the \emph{naïve} approach in \cite{trutman2022globally}). The average solve-time over 1000 samples was $\qty{0.17}{\second}$ for \emph{Gurobi}, $\qty{4.5}{\second}$ for \emph{SCIP}, and $\qty{3.3}{\second}$ for the \emph{naïve} SOS. 

Finally, we tested the SOS-based method without symbolic reduction on \href{https://www.asc-csa.gc.ca/eng/iss/canadarm2/about.asp}{Canadarm2}. While the average runtime was $\qty{6.3}{\second}$, the success rate was only $31 \%$ out of 1000 feasible instances. The average runtime for QCQP with \emph{Gurobi} was~$1.8\,\si{\second}$.

\subsection{\emph{iCub} Right Arm and Torso (7--10 DOF)}
We also demonstrate that our technique scales to realistic instances with up to 10 degrees of freedom. Due to the low range of motion of the iCub's arms, it can be beneficial to include the torso in the kinematic chain. We start with the 7-DOF right arm and progressively add torso joints up to the full 10-DOF chain. We are aware of no suitable baseline for this class of problems. 

\begin{table}
    \renewcommand{\arraystretch}{1.3}
    \caption{The QCQP runtime statistics (mean, first and third quartile) and the mean positional and rotational errors of the final solution computed from 1000 IK instances of iCub (7--10 DOF).}
    \label{icubtable}
    \centering
    \begin{tabular}{r||rrr|r|r}
    \hline
    & \multicolumn{3}{c|}{\bfseries{Time}} & \bfseries{Loc. Err.} & \bfseries{Rot. Err.}\\
    \bfseries{Set} & \shortstack{\bfseries{Avg.} \\ $(\mathbf{\si{\second}})$} & \shortstack{\bfseries{Q1} \\ $(\mathbf{\si{\second}})$} & \shortstack{\bfseries{Q3} \\ $(\mathbf{\si{\second}})$} & \shortstack{\bfseries{Avg.} \\ $(\mathbf{\si{\micro\meter}})$} & \shortstack{\bfseries{Avg.} \\ $(\mathbf{\si{\micro\radian}})$} \\
    \hline
    { 7 DOF} &  0.2 &  0.1 &  0.2 & 0.2 & 1.0 \\
    { 8 DOF} &  0.8 &  0.6 &  0.9 & 0.3 & 1.4 \\
    { 9 DOF} &  6.5 &  2.8 &  9.8 & 0.5 & 2.1 \\
    {10 DOF} & 77.6 & 28.0 & 90.4 & 0.8 & 2.9 \\
    \hline
    \end{tabular}
\end{table}

We also checked the scalability by running the 10-DOF instances on 126 threads of \emph{AMD EPYC 7543} processor, which resulted in a mean runtime of $\qty{10.1}{\second}$ over 100 samples.

\subsection{Quality of Local Solutions}
In case of the \emph{iCub} test sets, the objective value of the local solutions found by \emph{IPopt} was often either globally optimal already, regardless of the DOF. For poses from the 10-DOF set, \emph{Gurobi} was able to use the local solutions or find nearby feasible solution in $93\%$ of cases.

The median relative suboptimality of the local solutions was $-0.07\,\%$ for $10\,000$ feasible poses of the 8-DOF \emph{iCub} kinematic chain. Compare this to the \emph{KUKA iiwa}, where the median relative distance was $-14.8\,\%$.

\begin{figure}[tpb]
    \includegraphics[width=\linewidth]{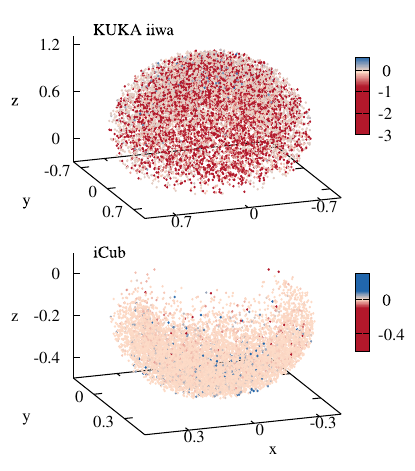}
    \caption{Difference between the objective achieved by \emph{IPopt} from the actual solution over $10\,000$ feasible poses showing that local solutions are often nearly optimal for the \emph{iCub} 8 DOF set, but not for \emph{KUKA iiwa}.}
    \label{fig:relsubopt}
\end{figure}

\subsection{Infeasible Poses}
Deciding that a kinematic chain is incapable of achieving a given pose may be faster or slower on average than computing the optimal angles for a feasible pose. We show this behavior on poses generated by computing the forward kinematics of randomly sampled joint angles without considering joint limits. 

Out of a set of 1000 poses generated for the KUKA manipulator, 345 were infeasible, and the average time to detect infeasibility was $\qty{1.1}{\second}$ compared to $\qty{0.44}{\second}$ to solve a feasible pose. The infeasibility of poses for iCub's arm (7~DOF) was detected in $\qty{0.02}{\second}$ on average over 980 samples.

\begin{figure}[tpb]
  \includegraphics[width=\linewidth]{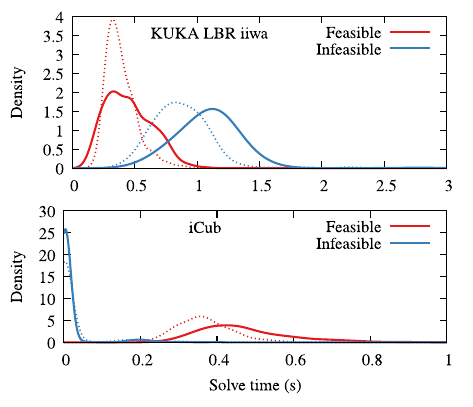}
    \caption{Kernel density estimation of solve times of feasible/infeasible poses using $\lift_\text{A}$. The performance of $\lift_\text{M}$ is overlaid as a dotted line.}
\end{figure}

\section{CONCLUSIONS}

We have demonstrated that our QCQP-based method finds globally optimal solutions to IK problems without symbolic reduction in exact arithmetic. Our results agree with those in \cite{trutman2022globally} for cases where IK instances can be set to solve the same objective.
 To the best of our knowledge, our technique is unique in the ability to globally solve IK instances of generic manipulators with eight or more degrees of freedom. 

Although our method does not apply to real-time control, additional pre-processing and appropriate heuristics may make such a use case possible. For example, we could use modern real-time IK solvers based on convex QP as runtime heuristics or generate initial feasible solutions to warm-start the search. 
While running the experiments, we observed that a significant fraction of the solution time for \emph{Gurobi} was spent proving that the solution was optimal. Including outer approximations in original variables as additional constraints or cuts will likely result in faster solve times. Similarly, the performance of \emph{SCIP} will likely improve with higher-quality initial solutions. It may also be possible to accelerate the detection of infeasible poses using the outer approximation technique described by Dai \textit{et al.} \cite{Dai2019-uf}.

Our proof-of-concept implementation is available on GitHub at \url{https://github.com/votroto/IK.jl}.

\section*{ACKNOWLEDGMENT}

We would like to thank Didier Henrion (LAAS-CNRS, University of Toulouse) and Jakub Mareček (Czech Technical University in Prague) for their valuable comments.


\bibliographystyle{IEEEtran}
\bibliography{bibliography}

\end{document}